\documentclass{article} 
\usepackage{iclr2026_conference,times}

\usepackage{amsmath,amsfonts,bm}

\def\eqref#1{equation~\ref{#1}}

\def\1{\bm{1}}

\DeclareMathAlphabet{\mathsfit}{\encodingdefault}{\sfdefault}{m}{sl}
\SetMathAlphabet{\mathsfit}{bold}{\encodingdefault}{\sfdefault}{bx}{n}

\usepackage{hyperref}
\usepackage{url}

\usepackage{amsmath}
\usepackage{microtype}
\usepackage{url}
\usepackage{booktabs}
\usepackage{graphicx}
\usepackage{lineno}
\usepackage{multirow}
\usepackage{algorithm}
\usepackage{algorithmic}
\usepackage{listings}
\usepackage{amsmath}
\usepackage{amssymb}
\usepackage{graphicx,scalerel}
\usepackage{xcolor}
\usepackage{cancel}
\usepackage{svg}
\newcommand\wh[1]{\hstretch{2}{\hat{\hstretch{.5}{#1}}}}

\title{Optimal Singular Damage: Efficient LLM\\ Inference in Low Storage Regimes}

\author{Mohammadsajad Alipour, Mohammad Mohammadi Amiri\\ Department of Computer Science\\ Rensselaer Polytechnic Institute\\ Troy, NY 12180, USA \\ \texttt{\{alipom,mamiri\}@rpi.edu} \\ }

\iclrfinalcopy 
\begin{document}

\maketitle

\begin{abstract}
Large language models (LLMs) are increasingly prevalent across diverse applications. However, their enormous size limits storage and processing capabilities to a few well-resourced stakeholders. As a result, most applications rely on pre-trained LLMs, fine-tuned for specific tasks. However, even storing the fine-tuned versions of these models remains a significant challenge due to the wide range of tasks they address. Recently, studies show that fine-tuning these models primarily affects a small fraction of parameters, highlighting the need for more efficient storage of fine-tuned models. This paper focuses on efficient storage of parameter updates in pre-trained models after fine-tuning. To address this challenge, we leverage the observation that fine-tuning updates are both low-rank and sparse, which can be utilized for storage efficiency. However, using only low-rank approximation or sparsification may discard critical singular components that enhance model expressivity. We first observe that given the same memory budget, sparsified low-rank approximations with larger ranks outperform standard low-rank approximations with smaller ranks. Building on this, we propose our method, optimal singular damage, that selectively sparsifies low-rank approximated updates by leveraging the interleaved importance of singular vectors, ensuring that the most impactful components are retained. We demonstrate through extensive experiments that our proposed methods lead to significant storage efficiency and superior accuracy within the same memory budget compared to employing the low-rank approximation or sparsification individually.
\end{abstract}

\section{Introduction}

Pre-trained language models have transformed the field of natural language processing by demonstrating impressive capabilities across a wide range of tasks.
These models are trained on a massive amount of text data, enabling them to learn complex language patterns and representations. However, their full potential is often realized through fine-tuning, a process where these pretrained models are further adapted to specific tasks using additional, often smaller datasets \citep{howard-ruder-2018-universal}.

Fine-tuning typically results in small changes to the original model, allowing for effective storage and significant memory savings \citep{hu2021lora}. Leveraging the small nature of fine-tuning updates, we can achieve significant memory efficiency with minimal performance loss.
Given the growing number of fine-tuned models derived from the same pre-trained backbone, efficiently storing these models is essential for scalable and sustainable deployment. Compressing model updates not only reduces memory and storage costs, but also facilitates faster transmission, easier updates across distributed systems, and lower energy consumption—factors that are critical in modern machine learning infrastructures such as cloud services, IoT ecosystems, and federated learning environments. 

While this compression does not reduce the one-time compute cost of fine-tuning, it brings substantial \emph{practical benefits} in several increasingly common scenarios \citep{yao2023deltazip}: \textbf{Federated or multi-task model hubs}, where hundreds of model updates must be stored and served for diverse user populations or downstream tasks; \textbf{Edge and on-device deployments}, where memory constraints make it infeasible to store full-size or uncompressed models; \textbf{Continual learning and versioning systems}, where model snapshots accumulate over time and quickly outgrow storage budgets. In such contexts, compressed model updates are key to enabling efficient storage, rapid model switching, and scalable deployment of personalized or task-specific models.

While parameter-efficient fine-tuning (PEFT) methods, such as Low-Rank Adaptation (LoRA) \citep{hu2021lora}, try to make the fine-tuning process efficient, there are cases where we have already fine-tuned the pre-trained model using classic full fine-tuning.
\citet{ryu2023efficient} and \citet{ping24} approach this problem from another perspective. They indicate that while PEFT methods like LoRA can maintain a certain level of accuracy, achieving accuracy close to that of full fine-tuning might be necessary in some cases. In other words, the amount of performance that we lose for efficiency in LoRA might be unacceptable in certain cases. So, they assume that we have already done the classic full fine-tuning and got our fine-tuned model. 
They tried to make the storage of fine-tuned models efficient by compressing the final difference between a fine-tuned model and its pre-trained version, referred to as the model updates, using the observation that this difference is a low-rank matrix \citep{hu2021lora}. They take advantage of this observation by performing a low-rank approximation method, truncated singular value decomposition (SVD), to store the model updates with much fewer parameters. Although many studies have explored low-rank approximation and sparsification of model updates, they have not effectively studied their great potential when they are combined, especially in low-storage scenarios.

In this paper, we aim to optimize the performance-efficiency trade-off by compressing the model updates, i.e., the difference between fine-tuned model and its corresponding pre-trained model. Our approach capitalizes on two key properties of these updates: their low-rank structure and inherent sparsity. By exploiting these characteristics, we achieve substantial reductions in storage requirements without compromising model performance. Specifically, we first obtain a low rank approximation of the original updates and leverage the inherent sparsity in this low-rank representation to further improve storage efficiency. This is accomplished by incorporating the relevant importance of each model parameter, ensuring that critical information is preserved while minimizing storage overhead. We propose a novel framework for sparsified low-rank approximation of model updates that outperforms approaches relying solely on low-rank approximation or sparsification, particularly in extreme compression regimes. 

Our method is based on the insight that combining these two techniques can preserve critical information often lost when they are applied independently. Instead of directly applying a strict low-rank approximation via truncated SVD with a fixed rank budget, we first perform truncated SVD with a relaxed rank, capturing a broader set of singular vectors. This ensures a richer representation of the model updates. To adhere to the memory budget, we then apply a structured sparsification strategy to this low-rank approximation, selectively retaining the most important parameters. This two-step process\textemdash low-rank approximation followed by sparsification\textemdash provides a greater flexibility in preserving essential information encoded in singular values, which is typically discarded in conventional one-step low-rank approximations due to their rigid rank constraints.

To guide the sparsification process, we introduce our main method \textbf{O}ptimal \textbf{S}ingular \textbf{D}amage (OSD), which leverages the interleaved importance of singular values by estimating the significance of model parameters using a first-order Taylor approximation of the loss function. This allows us to prioritize and retain the most impactful parameters while maintaining the overall storage budget. By unifying the complementary strengths of low-rank approximation and sparsification, our framework achieves superior performance compared to applying either technique in isolation. Additionally, our method offers unique advantages: it enables multiple possible compressed versions of parameters within the same storage budget, enhancing flexibility.
Our main contributions are summarized as follows:
\begin{itemize}
   \itemsep0em
   \item We propose a two-step, fast and flexible method for compressing model updates, the change in the pre-trained model after fine-tuning.
   \item We introduce an interleaved importance-aware sparsification method that preserves crucial singular vectors while maintaining a fixed storage budget.
   \item Our proposed framework allows the users to have and compare multiple possible compressed fine-tuned models within a desired budget.
   \item Our method achieves superior performance when there is a fixed budget of storage compared to performing truncated SVD and sparsification individually, particularly for low storage budget regimes.
\end{itemize}

\section{Related Work}
In recent years, several works have been proposed for efficient storage of model updates by leveraging sparsity, low-rank approximations and quantization. One such approach, DeltaZip \citep{yao2023deltazip} compresses model updates using structured 2:4 sparsity \citep{zhou2021learning,hubara2021accelerated} followed by quantization of the sparsed matrix to squeeze values into a smaller bit-width format. BitDelta \citep{liu2024bitdelta} quantizes model updates down to 1 bit. Then, they try to find the best rescaling factor for each approximated model update matrix that minimizes the original model update.

Beyond direct compression, several methods have been proposed on dynamic use of low-rank approximation and sparsification \citep{pmlr-v202-li23ap,zhang2025oatsoutlierawarepruningsparse,boza2025two}. One line of work \citep{pmlr-v202-li23ap,zhang2025oatsoutlierawarepruningsparse} involves applying the low-rank approximation to model weights and then adding a sparsified matrix by optimizing the sparsification of the reconstruction error between the low-rank approximation and the original weights. This process typically induces some form of iterative optimization overhead.  

These approaches often apply independent low-rank and sparsity constraints without considering their interplay in SVD decompositions. Moreover, sparsification of two matrices obtained through factorization of a matrix by decomposing it into the multiplication of two matrices has been shown to be more effective than applying sparsity to a single matrix \citep{boza2025two}, which provides insight into the success of low-rank approximation techniques. This also aligns with our claim regarding the benefits of enforcing sparsity in more relaxed compressed matrices compared to having a stricter compression ratio.

For low-rank approximation, several advanced SVD truncation methods have been proposed. Data-aware techniques that incorporate data whitening, layer sensitivities or loss function feedback to ensure a direct mapping between singular values and compression loss have been proposed by \citet{wang2025svdllm,wang-etal-2025-svd-llm},\citet{yuan2025asvd} and \citet{NEURIPS2021_f56de5efdrone}. Another approach, introduced by \citet{hsu2022languageW}, multiplies Fisher information matrix by truncated SVD of weights and tries to minimize the approximation error caused by the low-rank approximation. However, they neglect the interleaved importance of decomposed matrices in SVD and their potential for sparsification.

We highlight that our approach is orthogonal to these approaches, and one can use more complex SVD truncation methods or more complex sparsification methods in the components of our framework.

\section{Problem Formulation and Preliminaries}

We consider the problem of compressing the model updates for deployment during the inference. We consider layer-wise operations, where $W_f^l$ and $W_p^l$ are the parameters of the $l$-th layer in the fine-tuned and pre-trained models $W_f$ and $W_p$, respectively, with ${\Delta W}^l=W_f^l-W_p^l$ denoting the model updates for the $l$-th layer. 

For simplicity of notation, we assume that the model parameters have the same dimension across the layers, i.e., ${\Delta W}^l \in  \mathbb{R}^{n\times d}$, while our problem can be readily extended to the general case. Also, we denote the total number of layers in the model by $L$.
We define a compression function $\nu$ that is applied to the model updates $\Delta W^l$, i.e., $\nu(\Delta W^l)$ denotes the compressed form of $\Delta W^l$. We denote the memory (in bits) required for the compressed form of $\Delta W^l$ and its reconstruction by $\mu^l$. We denote the reconstruction or decompression function by $\rho$, where $\rho(\nu(\Delta W^l))$ provides an estimation of the original model updates after compression with $\nu$. Thus, the resulting model parameters after compression for layer $l$ is ${\wh{W}_f^l}=W_p^l+\rho(\nu(\Delta W^l))$ . Given a utility $P$ (e.g. accuracy), the objective is:
\begin{align}
    &\max_{\nu,\rho} P(\{\wh{W}_f^l\}_l) \\
    &{s.t.} \quad {{\mu^l}\leq B, \quad \forall l \in \{1,2,...,L\} }
\end{align}
That is, the goal is to find a compression function $\nu(\cdot)$ and a reconstruction function $\rho(\cdot)$ that maximizes the utility while maintaining a memory budget $B$ for storing the compressed model updates per-layer.

For the special case of no compression, i.e., when $\nu(\Delta W^l)=\Delta W^l$, the memory required to store $\nu(\Delta W^l)$ is given by $\nu^l=32 n d$ bits, where we represent each real number using a 32-bit floating-point format. Next, we present two baseline approaches to address the above problem: truncated SVD and sparsification.
\subsection{Low-rank Approximation via Truncated SVD}
Here, we present truncated SVD for compression of the model updates and analyze the resulting memory requirements \citep{ryu2023efficient}. In this approach, we compress $\Delta W^l$ by applying truncated SVD with rank $k \leq \min(n,d)$ to the model updates $\Delta W^l$ and approximate it with ${U_k} {\Sigma_k} {{V_k}^T}$, where ${U_k} \in \mathbb{R}^{n \times k}$, ${\Sigma_k} \in \mathbb{R}^{k \times k}$, and $V_k \in \mathbb{R}^{d \times k}$. 
We store $U_k \Sigma_k=\Delta W^lV_k \in \mathbb{R}^{n \times k}$ as the compressed form of $\Delta W^l$ and ${V_k^T} \in \mathbb{R}^{k \times d}$ used for reconstruction, and during inference, we decompress the model updates and estimate the original model parameters as $\wh{W}_f^l=W_p^l+U_k \Sigma_k {V_k}^T$. Accordingly, the required memory is given by:
\begin{align}\label{analysisSVD}
    \mu^l=  32(kn + kd) =   32 k({n+d}),
\end{align} 
where $32 kn$ bits and $32 kd$ bits are required to store $U_k\Sigma_k$ and ${V_k}$, respectively. We refer to this method as $\text{truncSVD}$ with rank $k$.

\subsection{Sparsification}
Preserving only a small fraction of model updates through hard thresholding can still maintain a certain amount of the performance of fine-tuned models, as demonstrated by \citet{isik2023gptzip,frantar2023sparsegpt,sun2024a}. We denote the sparse version of the matrix $\Delta W^l$ by top-$s(\Delta W^l)$, obtained by retaining only the $s$ entries with the largest absolute values.
To establish a rigorous evaluation of sparsification of model updates, we need to consider that, in order to store a sparse matrix top-$s(\Delta W^l)$ with $s$ non-zero entries, we need to store these non-zero values and their corresponding indices to locate them in the matrix. Thus, the number of stored bits is given by:
\begin{align}\label{analysisSparse}
    \mu^l= 32 s+\lceil \log_2(n d)\rceil s=(32+\lceil \log_2(n d)\rceil)s,
\end{align}
where $32s$ bits are required to store the $s$ non-zero values, and their indices are stored with $\lceil \log_2(n d)\rceil)s$ bits.
\section{Motivating Observation}\label{magnitude}
We observe that model updates exhibit high sparsity and we propose a balanced integration of low-rank compression and sparsification to maximize utility under a fixed memory budget. Specifically, we exploit the fact that truncated SVD with an increased rank budget $k=r+c$ (for some $c\in \mathbb{N}$) yields a more accurate approximation\textemdash retaining more singular values and reducing spectral norm error compared to a strict rank $k=r$ truncation. If the resulting low-rank approximation remains sparse, we can relax the rank constraint and allocate part of the memory budget to sparsification, further optimizing the trade-off between compression and fidelity. Here, we analyze the memory requirements of the $\text{truncSVD}$ method and our proposed approach combining low-rank approximation with sparsification. By equating their memory budgets, we derive the necessary sparsity level for our integrated approach to maintain equivalent storage costs. We note that the decomposed matrices $U_k\Sigma_k$ and $V_k$ can undergo independent sparsification, with sparsity levels denoted as $s_u$ and $s_v$, respectively.

For $k=r$, from Eq.~\ref{analysisSVD}, the memory required to store $U_r\Sigma_r$ and $V_r$ are given by $\mu^l_{u}=32rn$ and $\mu^l_v=32rd$, respectively. On the other hand, for the integrated approach with a low-rank compression of rank $k=r+c$, from Eq.~\ref{analysisSparse}, the sparse versions of $U_{r+c}\Sigma_{r+c}$ and $V_{r+c}$, with respective sparsity levels of $s_u$ and $s_v$, require a memory budget of $\mu^l_{u,s}=(32+\lceil \log_2(n (r+c))\rceil)s_u$ and $\mu^l_{v,s}=(32+\lceil \log_2(d (r+c))\rceil)s_v$, respectively. By setting $\mu^l_{u,s}=\mu^l_u$ and $\mu^l_{v,s}=\mu^l_v$ , for a fixed $r$ and $c$, the sparsity levels are given by:
\begin{align}
    \label{GU}
    s_u&=\lfloor \frac{32rn}{32+\lceil \log_2(n (r+c))\rceil}\rfloor, \\ \label{GV}
    s_v&=\lfloor \frac{32rd}{32+\lceil \log_2(d (r+c))\rceil}\rfloor.
\end{align}

\begin{figure}
    \centering
    \includegraphics[width=0.49\linewidth]{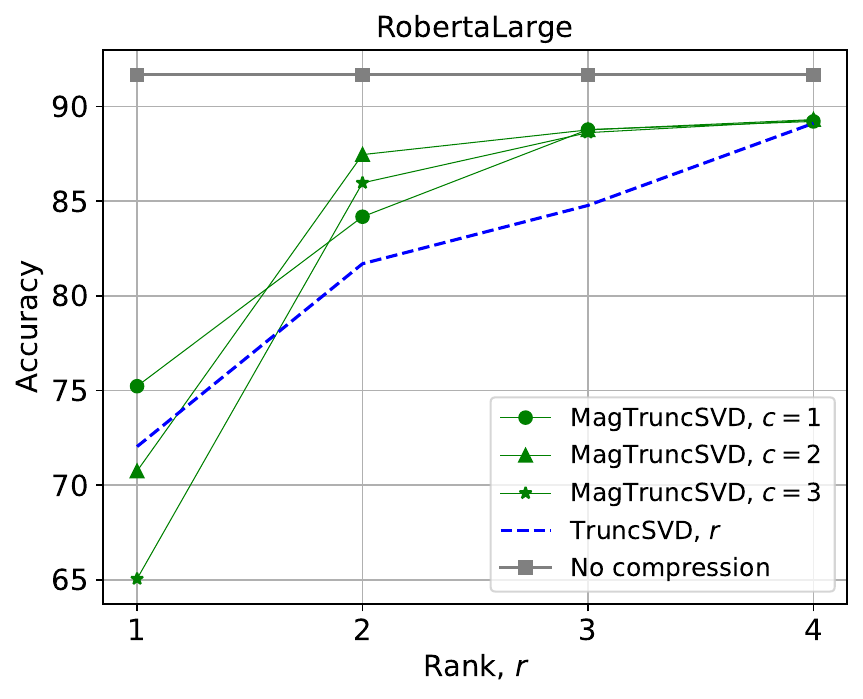}
    \includegraphics[width=0.49\linewidth]{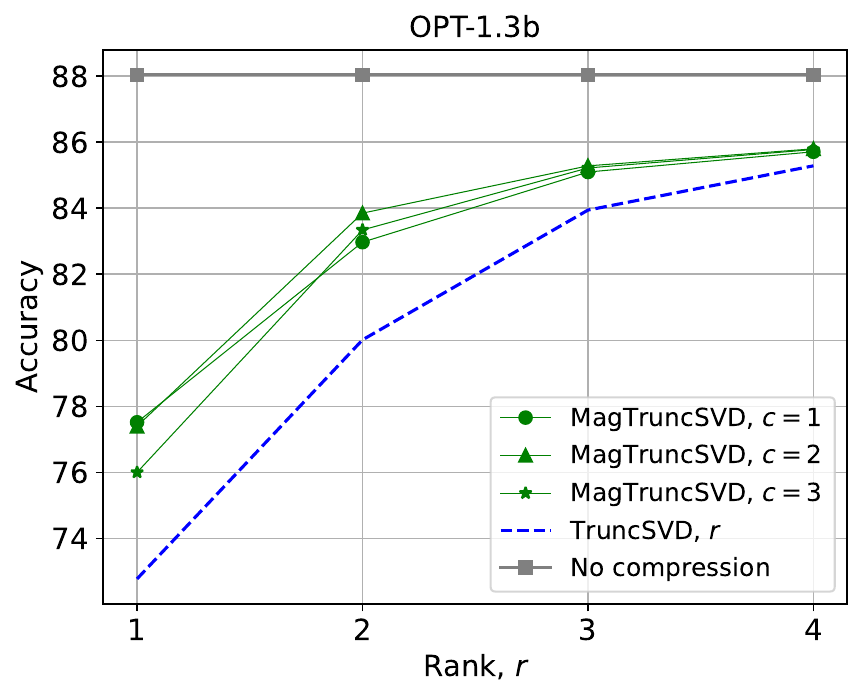}
    \caption{The average accuracy of the models RobertaLarge and OPT-1.3b on eight datasets in different natural language processing tasks including sentiment analysis, text classification, paraphrase detection and natural language inference, when we perform TruncSVD with rank $k=r$ and the proposed integrated approach (MagTruncSVD) combining TruncSVD with rank $k=r+c$ and sparsification for the same storage budget as TruncSVD with $k=r$.}
    \label{fig:init}
\end{figure}

Figure~\ref{fig:init} compares the performance between the conventional TruncSVD approach with strict rank $k=r$ and our proposed integrated approach, MagTruncSVD, combining TruncSVD with rank $k=r+c$ with sparsification under identical memory budgets. The evaluation demonstrates accuracy improvements for both the RobertaLarge \citep{DBLP:journals/corr/abs-1907-11692Roberta} and OPT-1.3b \citep{zhang2022opt} models. Notably, our hybrid approach MagTruncSVD consistently outperforms the standard TruncSVD for most tested $c$ values, with particularly significant gains in more constrained memory regimes.

These findings demonstrate that our integrated low-rank and sparsification approach provides a flexible generalization of conventional low-rank compression methods. By relaxing the rank constraint to $r+c$ rather than enforcing a strict rank-$r$ approximation, we create an adjustable parameter $c$ that can be optimized for enhanced model performance. This advantage stems from effectively exploiting the intrinsic dual structure of model updates, which exhibit both low-rank and sparse characteristics. Building upon this framework, we next introduce our main approach, referred to as OSD, that systematically combines low-rank approximation with a novel sparsification strategy. OSD advances our initial hybrid approach, MagTruncSVD, by developing a novel sparsification technique that dynamically captures the interleaved importance of factor matrices $U\Sigma$ and $V$.

\section{Optimal Singular Damage (OSD)}
Here we leverage the key insight from Figure~\ref{fig:init} to develop our method, OSD. That is, relaxing the rank constraint in low-rank approximation (allowing $k=r+c$) while strategically introducing sparsity in the factor matrices $U\Sigma$ and $V$ can outperform rank-$r$ truncation under some memory budget. Motivated by this, OSD integrates \textit{adaptive low-rank approximation} and \textit{structured sparsification} to compress model updates efficiently. The method proceeds in two phases:
\begin{enumerate}
    \item \textbf{Low-rank approximation}. Apply TruncSVD with rank $k=r+c$ to decompose the update $\Delta W^l$ into $U_k \Sigma_k$ and $V_k^T$.
    \item \textbf{Structured sparsification}. Sparsify the factor matrices $U_k\Sigma_k$ and $V_k^T$ by maintaining $s_u$ and $s_v$ non-zero entries to $\text{sparse}_{s_u}(U_k\Sigma_k)$ and $\text{sparse}_{s_v}(V_k^T)$, respectively, preserving only the most salient parameters.
\end{enumerate}

During the inference, the compressed update is reconstructed as $\rho ( \nu (\Delta W^l) )=\text{sparse}_{s_u}(U_k\Sigma_k)\cdot \text{sparse}_{s_v}(V_k^T)$ and the model estimate is given by:
\begin{align*}
    \wh{W}_f^l=W_p^l+\text{sparse}_{s_u}(U_k\Sigma_k) \cdot \text{sparse}_{s_v}(V_k^T).
\end{align*}
Next, we present the memory analysis for OSD and introduce our novel sparsification technique that leverages the interleaved importance of factor matrices.
\subsection{Memory analysis in OSD}
The total memory cost of OSD is given by:
\begin{align*}
    \mu^l&=32(s_u+s_v)+\lceil \log_2(n (r+c))\rceil s_u\\&\quad+\lceil \log_2(d(r+c))\rceil s_v,
\end{align*}
where $s_u$ and $s_v$ denote the sparsity levels for sparsifying $U_k\Sigma_k$ and $V_k$, respectively. For a fixed target rank $r$, OSD dynamically adjusts the rank relaxation $c$ and the sparsity levels $s_u$ and $s_v$ to ensure a Pareto-optimal performance for a fixed memory budget.

\subsection{Importance-aware sparsification}
While magnitude-based sparsification is computationally efficient, it suffers from a critical limitation: the magnitude of the parameter update after fine-tuning may not necessarily reflect its actual importance to model performance \citep{NIPS1989_6c9882bb}. To address this, we propose a novel \textit{importance-aware sparsification} method that uses gradient-based metrics to assess each parameter's contribution to task performance and jointly optimizes sparsification across factor matrices $U$ and $V$ while incorporating parameters' importance.

We compute parameter importance using a first-order Taylor approximation, evaluating how removing each individual parameter affects model performance \citep{molchanov2019importance}. Specifically, we obtain a parameter importance matrix $Z^l \in \mathbb{R}^{n\times d}$, corresponding to $W_f^l$, for each layer $l$ as follows:
\begin{align*}
    {Z^l}[i,j]=|\frac{\partial{\ell(D,W_f)}}{\partial W_f^l[i,j]}  W_f^l[i,j]| ,
\end{align*}
where $\ell(D,W_f)$ is the loss function on a small validation set $D$ and $\frac{\partial{\ell(D,W_f)}}{\partial W_f^l[i,j]}$ is the ($i,j$)-th entry of the gradient of the loss function. While our initial method, MagTruncSVD, is entirely data-independent, this method necessitates a small task-specific validation set. 

We now formalize the integration of parameter importance into our sparsification framework. Given TruncSVD with rank $k=r+c$, let $U'=U_k\Sigma_k \in \mathbb{R}^{n \times k}$ and $V'=V_k^T \in \mathbb{R}^{k\times d}$ denote the factor matrices. Therefore, without any sparsification, TruncSVD recovers $W':=U'V' \in \mathbb{R}^{n\times d}$. Sparsifying individual elements in $U'$ and $V'$ propagates structured changes to $W'$. Specifically setting ($i,j$)-th entry of $U'$ to zero, i.e., $U'[i,j]=0$, affects only the $i$-th row of $W'$. Let $W'_{u_{i,j}}$ be the perturbed approximation after setting $U'[i,j]=0$. The L1 norm error between $W'$ and $W'_{u_{ij}}$ is given by:
\begin{align}
    \lvert\lvert W'-W'_{u_{i,j}}\lvert\lvert_1&=\sum_{t=1}^{d} |W'[i,t]-W'_{u_{i,j}}[i,t]|
    =\sum_{t=1}^{d}|U'[i,j] V'[j,t]|.
\end{align}

This measures the local impact of sparsification. To extend this to model performance, we incorporate precomputed importance score $Z^l[i,t]$ for each parameter $W'[i,t]$, defining a sensitivity matrix $Q_{U'} \in \mathbb{R}^{n\times k}$ where:
\begin{align}\label{qu}
    Q_{U'}[i,j]&=\sum_{t=1}^{d} Z^l[i,t]|W'[i,t]-W'_{u_{i,j}}[i,t]|=\sum_{t=1}^{d}Z^l[i,t]|U'[i,j] V'[j,t]|.
\end{align}
Each entry $Q_{U'}[i,j]$ captures the performance-aware cost of zeroing $U'[i,j]$, combining both approximation error and parameter importance. Similarly, we define the sensitivity matrix $Q_{V'} \in \mathbb{R}^{k \times d}$ which captures the performance-aware cost of zeroing each entry of $V'$ as follows:
\begin{align}\label{qv}
    Q_{V'}[i,j]&=\sum_{t=1}^{n} Z^l[t,j]|W'[t,j]-W'_{v_{i,j}}[t,j]|=\sum_{t=1}^{n}Z^l[t,j]|U'[t,i] V'[i,j]|,
\end{align}
where $W'_{v_{i,j}}$ is the perturbed approximation $U'V'$ by setting $V'[i,j]=0$. Finally, to satisfy our memory budget constraints while preserving the most critical parameters, we implement a coordinated sparsification strategy for the factor matrices $U'$ and $V'$. Given sparsity targets $s_u$ and $s_v$ specifying the number of non-zero elements in $U'$ and $V'$, respectively, we first construct a combined importance matrix $Q=[Q_{U'}\quad Q_{V'}]$ through concatenating $Q_{U'}$ and $Q_{V'}$. 

This unified representation enables direct comparison of all parameters' significance across both matrices. We then perform global sparsity by selecting indices corresponding to the top $s_u+s_v$ values in $Q$. The corresponding elements in $U'$ and $V'$ are retained while all others are set to zero, ensuring optimal preservation of model performance within the fixed storage budget (line 13-15, Algorithm \ref{osdalgo}). This joint optimization approach accounts for inter-matrix parameter dependencies that would be overlooked by independent sparsification of $U'$ and $V'$ (our initial approach in section~\ref{magnitude}).

Our sparsification approach comprehensively captures the interleaved importance of elements in both $U'$ and $V'$ by evaluating their combined impact on model performance. The method operates in a single pass without requiring iterative optimization, making it computationally efficient. The framework is modular by design\textemdash the parameter importance module $Z^l$ can be readily substituted with alternative importance metrics while maintaining the overall sparsification procedure.

\subsection{Model Estimation}
We now present the final component of the OSD algorithm: estimating the model updates and accordingly the fine-tuned model. We perform the above hybrid approach, combining TruncSVD with rank $k=r+c$ and importance-aware sparsification, for a range of $c$ values from 1 to $C$ (the details about $c$ value are provided in Appendix \ref{howi}). For each candidate $c$ value, we determine the sparsity levels $s_u$ and $s_v$ to adhere to the memory budget $B$. Let $U_c^l$ and $V_c^l$ denote the sparsified versions of $U'$ and $V'$, respectively, for layer $l$ at rank expansion $c$, sparsified using importance-aware sparsification. This yields the parameter estimate $W_p^l+U_{c}^lV_{c}^l$ for evaluation (line 17, Algorithm \ref{osdalgo}). After assessing all $C$ candidates, we select the optimal $c^*$ that maximizes the performance (line 19, Algorithm \ref{osdalgo}), producing the final estimated model as $ \wh{W}_f^l=W_p^l+U_{c^*}^lV_{c^*}^l\quad \forall l\in \{1,2,...,L\}.$ A complete implementation outline of OSD is provided in Algorithm~\ref{osdalgo}.

\begin{algorithm}[t]
\caption{Optimal Singular Damage (OSD)}
\begin{algorithmic}[1]\label{osdalgo}
\STATE Require: Memory budget $B$, rank $r$, $C$ as the maximum value of $c$ (more details in Appendix~\ref{howi})
\FOR{$c = 1$ to $C$}
    \FOR{$l = 1$ to $L$}
        \STATE $\Delta W^l=W_f^l-W_p^l$
        \STATE Perform TruncSVD with rank $k=r+c$
        \STATE $\Delta W^l=U_k\Sigma_kV_k^T$
        \STATE $U'=U_k\Sigma_k$
        \STATE $V'=V_k^T$
        \STATE Calculate $Q_{U'}$ with Eq.~\ref{qu}
        \STATE Calculate $Q_{V'}$ with Eq.~\ref{qv}
        \STATE $Q=[Q_{U'} \quad Q_{V'}]$
        \STATE Calculate $s_u$ and $s_v$ with Eq.~\ref{GU},\ref{GV}
        \STATE $\tau=\max_{s_u+s_v} (Q)$ (returning the $(s_u+s_v)$-th maximum value of $Q$)
        \STATE $U_c^l[i,j]=\begin{cases} 
                U'[i,j], & \text{if } U'[i,j] \geq \tau \\ 
            0,  &  \text{otherwise}
            \end{cases} \quad \forall i,j$
        \STATE $V_c^l[i,j]=\begin{cases} 
                V'[i,j], & \text{if } V'[i,j] \geq \tau \\ 
            0,  &  \text{otherwise}
            \end{cases} \quad \forall i,j$
    \ENDFOR
    \STATE Evaluate performance: $P[c]=P(\{W_p^l+U_c^lV_c^l\}_l)$
\ENDFOR
\STATE Find best $c$: $c^*=\arg\max\limits_{c} P[c]$
\STATE Store $U_{c^*}^l$ and $V_{c^*}^l$, $\forall l \in \{1,...,L\}$
\STATE Model estimate: $\wh{W}_f^l=W_p^l+U_{c^*}^lV_{c^*}^l \quad \forall l\in \{1,2,...,L\}$
\end{algorithmic}
\end{algorithm}

\section{Experiments}

\begin{table*}[h]
    \scriptsize
    \centering
   
    \begin{tabular}{c| c| c c c c c c c c | c}
        \toprule
        \textbf{Budget} & \textbf{Method} & \textbf{AGNEWS} & \textbf{QNLI} & \textbf{MNLI} &  \textbf{SST-2} &  \textbf{QQP} &  \textbf{IMDB} &  \textbf{BOOLQ} & \textbf{MRPC} & \textbf{Average}  \\
        \midrule
        \multirow{1}{*}{$\infty$}    & No compression &  95.30 & 94.69 & 90.16  & 96.44 & 92.14 & 92.31 & 82.97 & 89.39 & ---  \\
                               
        \midrule
        \multirow{3}{*}{$r=1$}    & TruncSVD & 76.18  & 54.42  &  31.82 & 93.69 & 71.18 & 91.57 & \textbf{75.41} & 82.03 &  72.04 \\
                               & MagTruncSVD & 81.57 & 77.85 & 35.65 & \textbf{93.81} & 75.58 & 91.57 & 70.00 & 82.78 & 76.10   \\     
                               & \textbf{OSD} &  \textbf{83.84} & \textbf{86.00} & \textbf{41.12}  & 93.58 &  \textbf{82.05} & \textbf{91.69} & 74.43  & \textbf{83.13} &   \textbf{79.48}   \\

        \midrule
        \multirow{3}{*}{$r=2$}    & TruncSVD & 70.84  & 90.99  &  56.5 & 94.38 & 83.61 & 91.74 & 78.62 & 86.9 & 81.70  \\
                               & MagTruncSVD & 90.99 & 92.51 & \textbf{80.5} & 94.95 & 86.31 & 92.02 & 78.75 & 86.14 &  87.77    \\     
                               & \textbf{OSD} &   \textbf{91.58} & \textbf{92.64} & 79.85 & \textbf{95.07} & \textbf{86.5} & \textbf{92.04} & \textbf{79.24} & \textbf{87.07} & \textbf{88.00}    \\

        \midrule
        \multirow{3}{*}{$r=3$}    & TruncSVD &   72.37 & 90.87 & 75.78 & 94.72 & 85.66 & 92.0 & 78.65 & 88.17 & 84.78  \\
                               & MagTruncSVD & 92.49 & 92.9 & \textbf{86.31} & 95.18 & 86.56 & \textbf{92.12} & 79.51 & 87.19 & 89.03 \\     
                               & \textbf{OSD} & \textbf{93.0} & \textbf{93.06} & 85.64 & \textbf{95.53} & \textbf{86.8} & 92.11 & \textbf{79.76} & \textbf{88.23} & \textbf{89.27}\\ 

        \midrule
        \multirow{3}{*}{$r=4$}    & TruncSVD &   92.32 & 91.76 & 86.91 & 95.18 & 86.77 & 92.1 & 79.48 & 88.41 & 89.12 \\
                               & MagTruncSVD & 93.24 & 92.95 & \textbf{87.41} & \textbf{95.64} & 86.63 & \textbf{92.17} & 79.88 & 87.83 &  89.47    \\     
                               & \textbf{OSD} &  \textbf{93.43} & \textbf{93.17} & 87.25 & 95.53 & \textbf{87.09} & 92.16 & \textbf{80.06} & \textbf{88.52} & \textbf{89.65}  \\ 
    
        \bottomrule
    \end{tabular}
     \caption{Results for RobertaLarge performance across standard TruncSVD, MagTruncSVD, and OSD compression methods, at different low-storage settings.}
    \label{tab:resultsRoberta}
\end{table*}

\begin{table*}[h]
    \scriptsize
    \centering
    \begin{tabular}{c| c| c c c c c c c c | c}
        \toprule
        \textbf{Budget} & \textbf{Method} & \textbf{AGNEWS} & \textbf{QNLI} & \textbf{MNLI} &  \textbf{SST-2} &  \textbf{QQP} &  \textbf{IMDB} &  \textbf{BOOLQ} & \textbf{MRPC} & \textbf{Average}  \\

        \midrule
        \multirow{1}{*}{$\infty$}    & No compression &  95.43 & 92.79 & 83.89 & 95.41 & 91.64 & 92.09 & 72.48 & 80.58 &  --- \\
        
        \midrule
        \multirow{3}{*}{$r=1$}    & TruncSVD &  57.55 & 86.33 & 37.2 & 93.58 & 82.69 & 87.65 & 63.67 & 73.57 & 72.78 \\
                               & MagTruncSVD &  81.87 & 86.34 & 56.41 & 94.38 & 84.1 & 88.11 & \textbf{66.09} & 76.93 & 79.28  \\     
                               & \textbf{OSD} &   \textbf{85.83} & \textbf{88.89} & \textbf{67.54} & \textbf{94.95} & \textbf{84.19} & \textbf{89.97} & 65.63 & \textbf{78.2} & \textbf{81.90}   \\

        \midrule
        \multirow{3}{*}{$r=2$}    & TruncSVD &   70.2 & 89.6 & 66.25 & 94.61 & 83.14 & 90.87 & 67.03 & 78.43 & 80.02 \\
                               & MagTruncSVD & 91.61 & 90.12 & 73.45 & \textbf{94.95} & \textbf{85.61} & 91.07 & \textbf{67.22} & 78.67 & 84.09 \\     
                               & \textbf{OSD} &   \textbf{92.5} & \textbf{90.54} & \textbf{78.56} & \textbf{94.95} & 85.43 & \textbf{91.5} & 66.36 & \textbf{79.25} & \textbf{84.89}    \\

        \midrule
        \multirow{3}{*}{$r=3$}    & TruncSVD &  86.72 & 90.3 & 76.82 & 94.84 & 85.2 & \textbf{91.74} & 66.64 & 79.3 & 83.94  \\
                               & MagTruncSVD & 93.04 & 90.74 & 79.07 & \textbf{95.07} & \textbf{86.3} & 91.46 & \textbf{68.29} & 79.36 & 85.42\\     
                               & \textbf{OSD} &  \textbf{93.3} & \textbf{91.01} & \textbf{81.23} & 94.95 & 86.08 & 91.67 & 67.28 & \textbf{79.42} & \textbf{85.62}   \\ 

        \midrule
        \multirow{3}{*}{$r=4$}    & TruncSVD &  93.08 & 90.87 & 79.87 & 94.61 & 85.86 & 91.72 & 66.91 & 79.36 & 85.29  \\
                               & MagTruncSVD & 93.29 & 91.23 & 81.32 & 95.07 & \textbf{86.57} & 91.6 & \textbf{68.32} & 79.83 & 85.90\\     
                               & \textbf{OSD} &   \textbf{93.51} & \textbf{91.25} & \textbf{82.84} & \textbf{95.18} & 86.45 & \textbf{91.77} & 67.92 & \textbf{79.94} & \textbf{86.11}    \\ 
          
        \bottomrule
    \end{tabular}
    \caption{Results for OPT-1.3b performance across standard TruncSVD, MagTruncSVD, and OSD compression methods, at different low-storage settings.}
    \label{tab:resultsForOPT}
\end{table*}

In this section, we evaluate and compare our initial hybrid approach MagTruncSVD and our main method OSD with TruncSVD as the baseline. For classification tasks, We conducted experiments on eight benchmark NLP tasks including SST-2 (sentiment classification), IMDB (sentiment analysis), AG News (news classification), MRPC (paraphrase identification), QQP (question paraphrase detection), QNLI (question-answering natural language inference(NLI)), MNLI (multi-genre NLI), and BoolQ (yes/no question answering), with two base models Roberta Large \citep{DBLP:journals/corr/abs-1907-11692Roberta} and OPT-1.3b \citep{zhang2022opt}. Additionally in Appendix \ref{gener}, for generative tasks, we conducted experiments with two base models, LLaMA-2-7B and LLaMA-2-13B \cite{touvron2023llama} on two benchmark datasets: GSM8K \cite{cobbe2021traininggsm8k}, multi-step numerical reasoning, and TruthfulQA \cite{lin2021truthfulqa}, a benchmark for measuring the truthfulness of model-generated answers.

\subsection{Results and Discussion}
Tables~\ref{tab:resultsRoberta} and \ref{tab:resultsForOPT} present the performance of fine-tuned RobertaLarge and OPT-1.3b models on the eight downstream tasks after compressing with standard TruncSVD, our initial method $\text{MagTruncSVD}$ based on magnitude sparsification and our main method OSD based on interleaved importance-aware sparsification. As can be seen in both models, on average, our methods have consistently outperform standard TruncSVD with rank $k=r$ within the same budget, by a significant margin. Moreover, OSD has surpassed our naive magnitude-based sparsification approach MagTruncSVD discussed in Section~\ref{magnitude}, supporting our claim that OSD effectively identifies and captures the interleaved importance of elements within the factor matrices $U_k\Sigma_k$ and $V_k$ based on their contribution to performance and approximation error. 

It is important to emphasize that, to ensure fairness, our approach stores a slightly fewer number of parameters than what is stored by the regular TruncSVD with rank $k=r$. This consideration stems from the fact that when storing a sparse matrix, it is not only the non-zero elements that need to be stored, but also their corresponding indices. Consequently, a matrix with larger dimensions would require more bits for storing these indices. By accounting for this factor, we ensured that our results remain equitable and reliable when compared to regular TruncSVD with rank $k=r$. On average, for the RobertaLarge model, OSD achieves a significant improvement of 7.44\% over standard TruncSVD in $r=1$. As $r$ increases, the performance gains continues, with a difference of 6.30\% at $r=2$, and 4.49\% at $r=3$. The smallest improvement is observed at $r=4$, where OSD outperforms TruncSVD by 0.53\%. For the larger model OPT-1.3b model, OSD delivers a substantial increase in accuracy by achieving 9.12\% and 4.87\% better performance compared to standard TruncSVD at $r=1$ and $r=2$, respectively. For $r=3$ and $r=4$, OSD has 1.68\% and 0.82\% improvement over standard TruncSVD, respectively. 

The experimental results demonstrate that the performance advantage of our method diminishes as $r$ increases. This trend occurs because when $r$ approaches the true rank of the model updates, $\text{rank}(\Delta W^l)\approx r$, the benefits of our relaxed rank approach (setting $k=r+c$) become marginal\textemdash standard TruncSVD with rank $k=r$ already achieves near-perfect approximation in this regime. We also observe that for certain sentiment analysis tasks (IMDB and SST-2), even highly compressed representations ($r=1$) preserve substantial performance. This explains why our improvements appear more modest on these tasks compared to others where low-rank approximations incur greater information loss.

\subsubsection{Computational Complexity of OSD}\label{complexityanalysis}
The computational complexity of OSD primarily consists of (1) computing the truncated SVD of ${\Delta W}^l\in \mathbb{R}^{n\times d}$ and (2) obtaining the importance matrices $Q_{U'}$ and $Q_{V'}$ across all $L$ layers. SVD of an $n \times d$ matrix incurs a computational complexity of $O(\min(nd^2,n^2d))$, where assuming that $n\ge d$, it becomes $O(nd^2)$. Calculating all elements in $Q_{U'}$ and $Q_{V'}$ incurs a complexity of $O(n(r+c)d)$. Moreover, the sorting operation on the concatenated importance matrices $Q_{U'}$ and $Q_{V'}$, demands $O((n+d)(r+c) log((n+d)(r+c)))$ time. Given that $r+c \ll \min(n,d)$ in practice, the dominant term for each layer becomes $O(nd^2)$. Consequently, the total complexity across all layers scales as $O(Lnd^2)$. We also provide an efficient method for computing the $Q_U$ and $Q_V$ values in Appendix~\ref{efficient_cal}. Importantly, OSD is performed offline and does not alter the model architecture or inference path, thereby introducing no additional inference-time latency, unlike methods that rely on auxiliary components.

\subsubsection{Limitations of Pure Sparsification Baselines}
In scenarios with extremely constrained memory budgets, relying solely on sparsification of $\Delta W^l$ proves substantially less effective than TruncSVD with rank $k=r$. This inefficiency arises because storing a high-dimensional sparse matrix under tight memory constraints necessitates an extreme reduction in the number of preserved parameters. To quantify this, consider the memory required for rank-$r$ TruncSVD in layer $l$, $\mu^l=32r(n+d)$. If we instead allocate the same budget $\mu^l$ to store a sparse representation of the full $n\times d$ matrix $\Delta W^l$, the maximum number of storable non-zero elements reduces to
$s=\lfloor{\frac{\mu^l}{32+\lceil \log_2(n d)\rceil}}\rfloor$. 
The logarithmic term\textemdash which accounts for the storage overhead of sparse matrix indices\textemdash becomes particularly the bottleneck in high dimensions. Consequently, pure sparsification preserves significantly less meaningful parameters than OSD under identical memory constraints, particularly in low-rank regimes where $r \ll \min(n,d)$.

\subsubsection{Effect of Parameter Importance in OSD}
\begin{figure}
    \centering
    \includegraphics[width=0.48\linewidth]{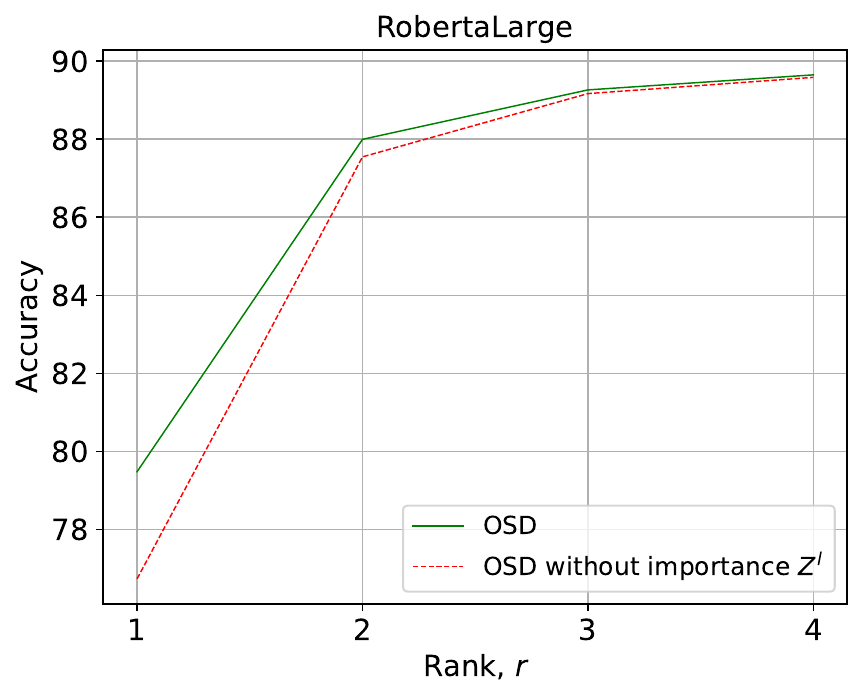}
    \includegraphics[width=0.48\linewidth]{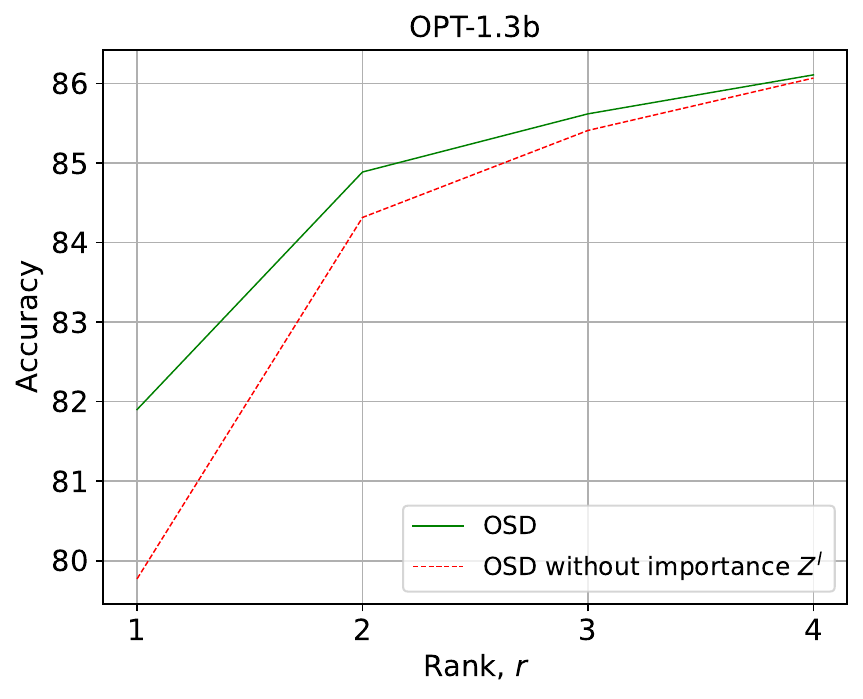}
    \caption{Effect of incorporating weight importance $Z^l$ into OSD and excluding it, on the model performance after approximation, for RobertaLarge and OPT-1.3b models.}
    \label{fig:ablation}
\end{figure}

We also conducted an ablation study to examine the practical impact of incorporating parameter importance module, denoted as $Z^l$, into our interleaved method OSD. The results of this study are depicted in Figure~\ref{fig:ablation}, which shows that the inclusion of parameter importance, i.e., multiplying $Z^l$ in Eq.~\ref{qu} and Eq.~\ref{qv}, leads to a noticeable improvement in the overall utility of the OSD method. This suggests that incorporating parameter importance plays a crucial role in enhancing the effectiveness of the OSD approach.

\section{Conclusion}
This work addresses the critical challenge of efficiently compressing the difference between fine-tuned and pre-trained models\textemdash a fundamental requirement for storage constraint deployment scenarios. We present a novel solution that significantly improves the performance-efficiency trade-off in extreme low-rank regimes. Our key insight reveals that relaxed low-rank approximations coupled with strategic sparsification outperform strict rank constraints, as they better preserve crucial update patterns with limited memory budget. Another innovative aspect of our proposed method lies in our importance-aware sparsification technique, which retains the most significant parameters across factor matrices by accounting for their interleaved contributions to the model performance. Experimental results demonstrate that our method achieves superior compression rates while maintaining competitive task performance compared to individual compression approaches. 

Beyond storage efficiency, our methods enable \emph{lightweight model sharing} across clients and \emph{lower memory overhead} for real-time adaptation---making them well-suited for deployment in bandwidth- or memory-constrained environments.
Future research directions include developing adaptive rank selection mechanisms to automatically optimize the rank-sparsity trade-off, and extending the framework to support PEFT under aggressive memory constraints. The principles underlying our approach may also benefit other large-scale machine learning scenarios where efficient model storage is paramount.

\bibliography{iclr2026_conference}
\bibliographystyle{iclr2026_conference}

\appendix
\section{Appendix}

\section{Optimal Rank Relaxation: The Role of c Value}\label{howi}
\begin{figure}[h]
    \centering
    \includegraphics[width=0.49\linewidth]{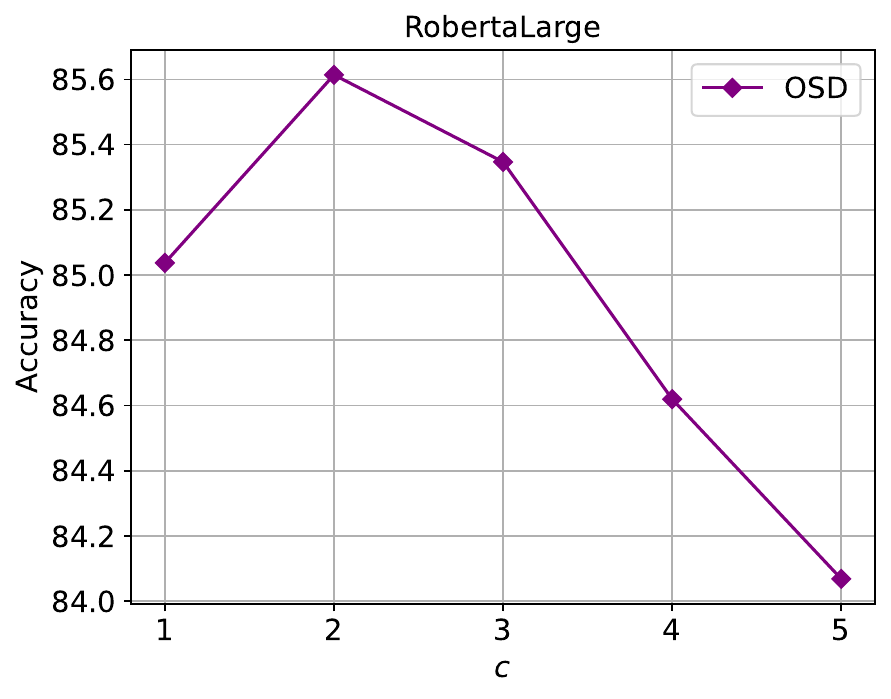}
    \includegraphics[width=0.49\linewidth]{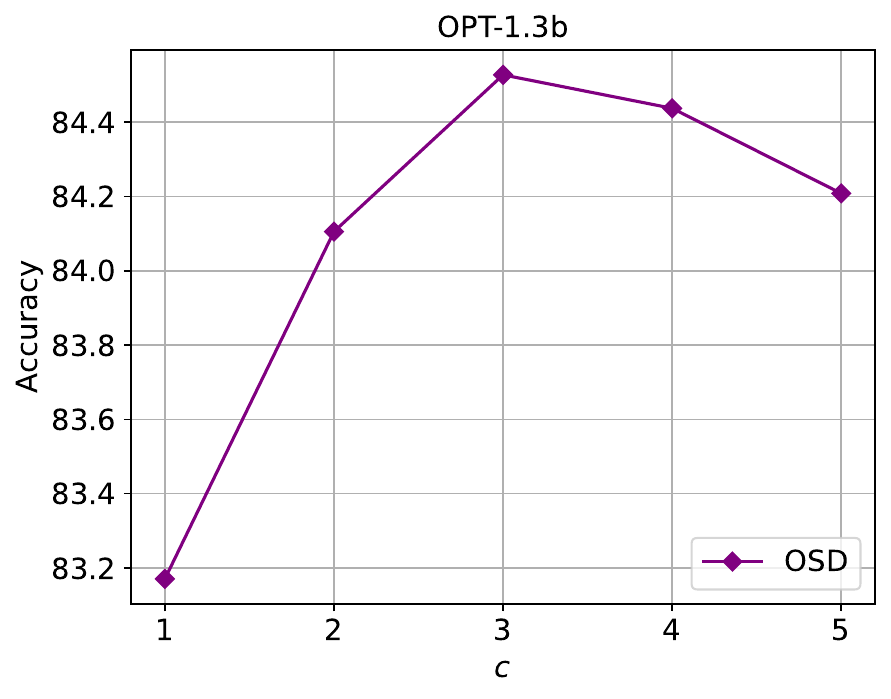}
    \caption{Average performance across all tasks and all values of $r$ ($1 \leq r \leq 4$) for different values of $c$ in the OSD algorithm, evaluated on the RobertaLarge and OPT-1.3b models.}
    \label{fig:cvalue}
\end{figure}

Our experiments reveal a characteristic bell-shaped performance curve as $c$ varies in the OSD method (Figure~\ref{fig:cvalue}). The initial performance improvement with increasing $c$ stems from capturing additional critical singular directions through the relaxed rank $k=r+c$. Even in the sparsified form, these extra components prove more valuable than their complete omission in the strict rank-$r$ truncation. However, beyond an optimal point, further increasing $c$ leads to performance degradation due to competing effects: while more singular components are retained, the available memory budget forces extreme sparsity in the factor matrices (as governed by Eq.~\ref{GU},\ref{GV}), ultimately causing information loss. 

This trade-off manifests because larger $c$ values reduce the number of storable non-zero parameters relative to standard TruncSVD with rank $k=r$. Through extensive empirical evaluation, we identify the sweet spot for $c$: optimal performance occurs within $1\leq c\leq 5$.

\section{Results for Generative tasks}\label{gener}

{Table~\ref{tab:resultsMath} presents the performance of fine-tuned base models—LLaMA-2-7B and LLaMA-2-13B—on GSM8K and TruthfulQA under varying compression budgets ($r = 1$ to $4$), comparing three methods: TruncSVD, our initial method MagTruncSVD, and our main method OSD.
Across both model sizes and datasets, we observe consistent trends highlighting the advantages of pruning higher rank truncated SVD of updates guided by magnitude (MagTruncSVD) and importance (OSD) over pure TruncSVD. 
Overall, these results demonstrate that: (1) naive compression via TruncSVD significantly degrades task performance; (2) MagTruncSVD provides a strong data-independent baseline; and (3) OSD achieves the best trade-off between compression and performance, particularly in more challenging reasoning tasks like GSM8K. Moreover, it is worth mentioning that we could not assess the full potential of our OSD method in this set of experiments due to the enormous size of models and consequently, their gradients for the importance matrix. Instead, we used an all-one matrix for the importance matrix $Z^l$, thereby focusing solely on the reconstruction error.

\begin{table*}
    \small
    \centering
    \begin{tabular}{c|c|cc|cc}
        \toprule
        \textbf{Budget} & \textbf{Method} 
        & \multicolumn{2}{c|}{\textbf{LLama-2-7B}} 
        & \multicolumn{2}{c}{\textbf{LLama-2-13B}} \\
        & & \textbf{GSM8K} & \textbf{TruthfulQA} 
          & \textbf{GSM8K} & \textbf{TruthfulQA} \\
        \midrule
        \multirow{3}{*}{$r=1$}    
            & TruncSVD        
            & 0.08 & 38.21 
            & 22.9 & 36.6 \\
            & MagTruncSVD     
            & 0.3 & 39.3
            & 24.79 & 38.09 \\
            & OSD* 
            & 0.38 & 39.2
            & 26.31 & 37.98 \\
        \midrule
        \multirow{3}{*}{$r=2$}    
            & TruncSVD        
            & 0.23 & 40.21 
            & 23.35 & 38.49 \\
            & MagTruncSVD     
            & 1.59 & 39.57
            & 26.08 & 38.24 \\
            & OSD* 
            & 1.67 & 39.76 
            & 25.7 & 38.6 \\
        \midrule
        \multirow{3}{*}{$r=3$}    
            & TruncSVD        
            & 2.81 & 40.26 
            & 23.81 & 39.02 \\
            & MagTruncSVD     
            & 7.13 & 39.92 
            & 25.63 & 38.48 \\
            & OSD*    
            & 6.07 & 40.64
            & 25.85 & 38.84 \\
        \midrule
        \multirow{3}{*}{$r=4$}    
            & TruncSVD        
            & 7.05 & 40.44 
            & 24.64 & 38.92 \\
            & MagTruncSVD     
            & 11.6 & 41.02 
            & 25.93 & 38.52 \\
            & OSD* 
            & 13.12 & 41.14 
            & 25.93 & 38.83 \\
        \bottomrule
    \end{tabular}
    \caption{Results for LLama-2-7B and LLama-2-13B performance across standard TruncSVD, MagTruncSVD, and OSD compression methods, at different low-storage settings.
    \\{*: For larger models, we used an all-ones importance matrix in OSD due to our computational limitations.}}
    \label{tab:resultsMath}
\end{table*}

\section{Efficient Calculation of $Q_{U'}$ and $Q_{V'}$}\label{efficient_cal}
Calculating each element in $Q_{U'}$ and $Q_{V'}$ could benefit from fast and parallel vector multiplication using the following code:
\begin{lstlisting}
    Q_U[i,j]=torch.sum(torch.abs((Z[i,:]*V[j,:]) * U[i,j])) 
    Q_V[i,j]=torch.sum(torch.abs((Z[:,j]*U[:,i]) * V[i,j]))
\end{lstlisting}

\end{document}